\documentclass[a4paper, 10 pt, conference]{ieeeconf}  

\IEEEoverridecommandlockouts                              
                                                          
\usepackage{xcolor}
\usepackage{amsfonts}
\usepackage{amsmath}
\usepackage{mathtools}
\usepackage{multirow}
\usepackage{cite}

\usepackage[pdftex]{graphicx}

\begin{document}
\title{Defining Traffic States using Spatio-temporal Traffic Graphs}

\author{Debaditya Roy, K. Naveen Kumar, C. Krishna Mohan
\thanks{D. Roy is with the Department of Transportation Systems Engineering, College of Science and Technology, Nihon University. K. N. Kumar and C. K. Mohan are with the Department of Computer Science and Engineering, Indian Institute of Technology Hyderabad.}
}

\maketitle

\begin{abstract}
Intersections are one of the main sources of congestion and hence, it is important to understand traffic behavior at intersections. Particularly, in developing countries with high vehicle density, mixed traffic type, and lane-less driving behavior, it is difficult to distinguish between congested and normal traffic behavior. In this work, we propose a way to understand the traffic state of smaller spatial regions at intersections using traffic graphs. The way these traffic graphs evolve over time reveals different traffic states - a) a congestion is forming (clumping), the congestion is dispersing (unclumping), or c) the traffic is flowing normally (neutral). We train a spatio-temporal deep network to identify these changes. Also, we introduce a large dataset called EyeonTraffic (EoT) containing 3 hours of aerial videos collected at 3 busy intersections in Ahmedabad, India. Our experiments on the EoT dataset show that the traffic graphs can help in correctly identifying congestion-prone behavior in different spatial regions of an intersection.
\end{abstract}

\IEEEpeerreviewmaketitle

\section{Introduction}
Intersections are a major cause of congestion in urban networks especially in the case of lane-less mixed traffic where a large number of smaller vehicles bunch together at stop signs in an irregular fashion. Determining the level of congestion at intersections is mainly performed in existing works by counting the number of vehicles \cite{congestioncnn}. While this is suitable for lane-based traffic, these methods are not applicable for high irregular traffic density with varying sizes of vehicles in developing countries like India. Furthermore, there is high propensity for lateral movements and low gap maintenance in such kind of traffic that may indicate congestion but actually a normal traffic state. 

\begin{figure}
 \centering
  \includegraphics[width=\linewidth,height=6cm]{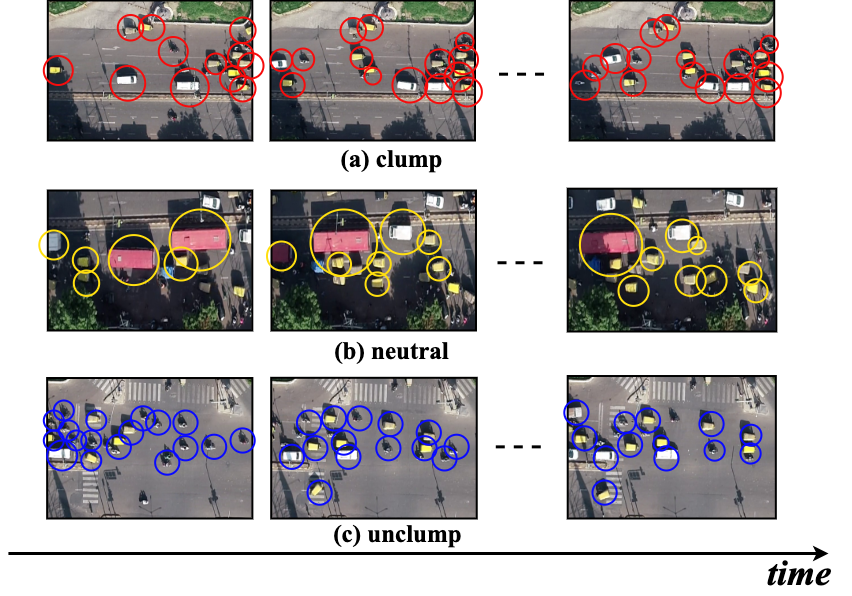}
  \caption{Different types of traffic states (a) clumping - the vehicles bunch together rapidly, (b) neutral - the vehicles maintain the same relative speed and gap, and (c) unclumping - the vehicles disperse away rapidly.  }
  \label{traffic_states}
\end{figure}

The aforementioned issues make it imperative to define a set of traffic states specifically for understanding mixed lane-less traffic. Hence, in this work, we propose a novel characterization of traffic states using traffic graphs. It is important to note that different spatial regions of an intersection are in different traffic states that are dependent on the traffic signals and the corresponding cycle length. So, the traffic states should be defined based on the changing interaction between vehicles over time as shown in Figure \ref{traffic_states}. The evolution of these interactions can be best represented using traffic graphs \cite{trafficgraph1} that change over time. Particularly, we observed that the traffic graph structure of a particular spatial region changes when too many vehicles congregate or disperse in a short span of time. The temporal pattern of changing neighbors is best represented using the adjacency matrix of the traffic graph. So, we propose to learn the spatio-temporal pattern of these adjacency matrices using a network consisting of Convolutional Neural Network (CNN) and Gated Recurrent Unit (GRU) units. 

Another challenge in investigating traffic states of lane-less traffic is that there is no dataset that contains vehicle trajectories at intersections. Most of the datasets are based on lane-based traffic like DETRAC \cite{detrac}, inD \cite{InD}, Interaction \cite{interaction}, and highD \cite{highd}. As the characteristics of lane-less mixed traffic are quite different, we introduce a new dataset called EyeonTraffic (EoT) that contains aerial videos of 3 intersections in Ahmedabad, India.  An hour of video is collected from each intersection and the corresponding vehicle trajectories are provided along with the spatio-temporal annotations of traffic states.   

\section{Related Work}
In literature, traffic state has been studied by measuring either traffic volume, density, or speed \cite{zhang2017understanding,congestioncnn}. Most of these works use either detection-based methods or aggregate approaches. Detection-based methods count vehicles by identifying and localizing them in video frames. Recently, faster recurrent convolution neural networks (FRCNNs) have been used for traffic density calculation \cite{frcnn}. However, FRCNNs have struggle with poor resolution videos and  heavy occlusion. In \cite{yolocounter}, the authors introduced multi-object tracking along with vehicle detection using the YOLOv3 architecture, to improve the performance of vehicle counting. A new architecture was proposed in \cite{hydraCNN} where vehicle counting and traffic density were obtained directly from CNN architecture called CountingCNN and HydraCNN, respectively. In \cite{zhang2017understanding}, a residual network to predict vehicle count and density simultaneously was proposed that could handle low frame rate and high occlusion in videos. 

Aggregation based approaches avoid the detection or segmentation of vehicles and analyze the entire image to estimate the overall traffic state. In \cite{gonccalves2012spatiotemporal}, traffic videos were categorized into different congestion types with the help of spatiotemporal Gabor filters. A simpler approach was proposed in \cite{lempitsky2010learning} where a linear transformation was applied on
each pixel feature to estimate the density of vehicles in an frame of traffic video. Furthermore, for lane-less traffic, a trajectory based method was introduced in \cite{raju2018application} where traffic states were assigned based on the flow of traffic, i.e., free-flowing, moderate flow, and  congested flow. Also, lateral and longitudinal movement of vehicles across lanes were considered to categorize traffic movement into six different traffic states. Finally, the gap following behavior between vehicles, the relative distance and velocities were also studied to understand the actual traffic state. A major bottleneck of this approach is that it requires a) manual annotation of each vehicle in the image to extract trajectories, b) exact measurement of vehicle speed and relative gap, and c) determining the maximum capacity of the road section under observation. As discussed earlier, these parameters require considerable effort and time to be computed in heavily occluded traffic.

\section{Proposed Approach}
\begin{figure}
 \centering
  \includegraphics[width=0.9\linewidth,height=22cm]{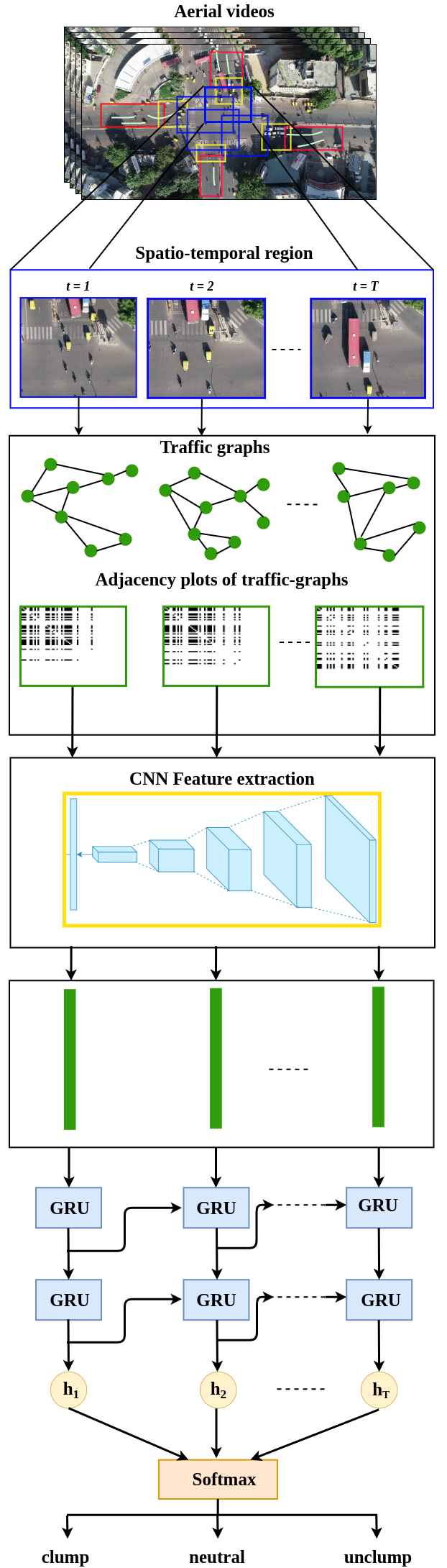}
  \caption{Block diagram of the proposed approach}
  \label{blockdiag}
  \vspace{-2mm}
\end{figure}

From the literature review, it can be observed that aggregation approaches provide a holistic interpretation of traffic state with manual calculation of the myriad of parameters needed. Hence, in this work, the traffic states that we propose are based on the movement of traffic density in a spatial region based on automated detection and tracking of vehicles. The block diagram of the entire approach is presented in Figure \ref{blockdiag}. At first the entire intersection is divided into smaller spatio-temporal regions based on the stop signs at each lane of the intersection. Next, the vehicles in each of these regions are detected using the RetinaNet architecture \cite{retinanet} and tracked using the DeepSort \cite{deepsort} tracker. The tracking is done for a specified time interval and the corresponding traffic graphs are extracted fro each time-step. Using a CNN, features are extracted from the adjacency matrices of these traffic graphs which are then sent as input to two GRU layers. Finally, the spatio-temporal information from the traffic graphs is classified into one of the three traffic states.  

\subsection{Traffic states using traffic graphs}
Traffic in a spatial region with $n$ road users can be represented using a traffic-graph $G_t$ for a particular time-step $t$. The spatial positions of the road users at that time-step are denoted by the set of vertices $V_t ={v_1^t, v_2^t, \cdots , v_n^t}$ and the relation between the road users is denoted a set of undirected, weighted edges, $E_t$. Any two road users can be connected through an edge if $d(v_i^t, v_j^t ) < \mu$, where $d(v_i^t, v_j^t )$ represents the Euclidean distance between the road-agents and $\mu$ is a threshold. After inspecting the aerial videos in the EyeonTraffic dataset, $\mu$ was chosen to be 10 meters to retain those interactions the affect the road users (based on the size of the road-users and the width of the road).

In order to determine the changing relationship of a road user with its neighbors, a symmetrical adjacency matrix $\textbf{A}_t \in \mathbb{R}^{n\times n}$ is calculated at every time-step for the corresponding traffic-graph $G_t$ as,
\begin{equation}
\mathbf{A}_t (i,j) =  \left\{\begin{matrix}
e^{-d(v_i^t,v_j^t)} & \text{if } d(v_i^t, v_j^t ) < \mu \text{ and } i \ne j\\ 
0 & \text{otherwise}.
\label{equ}
\end{matrix}\right.
\end{equation}

The distance function $e^{-d(v_i^t,v_j^t)}$ denotes the interactions between any two road users $v_i$ and $v_j$ at time $t$. Using an decaying exponential function leads road users that are far away being assigned a lower weight compared to road-agents in close proximity. This is in line with the observation that each road user pays more attention to other nearby road users in order to avoid traffic collisions. 

\begin{figure}
  \includegraphics[width=\linewidth]{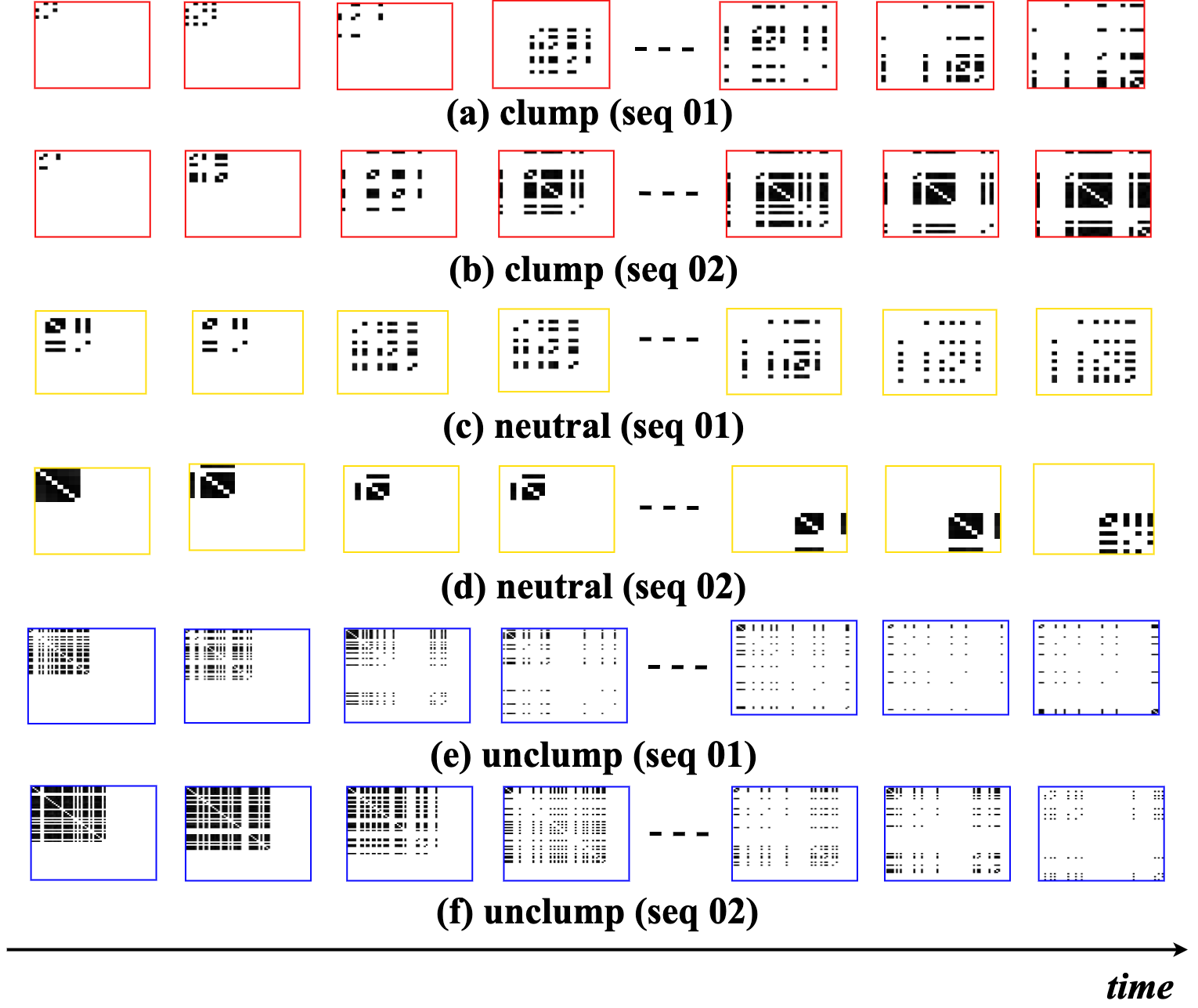}
  \caption{Evolution of adjacency matrices for different traffic states.}
  \label{adjacency}
  \vspace{-5mm}
\end{figure}

Observing the adjacency matrices allows us to categorize traffic movements into 3 distinct states - clumping, unclumping, and neutral. As shown in Figure  \ref{adjacency}(a) and (b), in case of \textit{clumping}, the adjacency matrix is sparse in the beginning as road users are far away from each other. As more road users converge towards the same spatial location i.e. the stop line of the intersection, the interactions between the road users increase leading to a more dense adjacency matrix. In case of \textit{neutral} state, the rate of change of adjacent road users is zero or very low. Since the number of interactions do not vary much over time, the density of the adjacency matrix remains similar over time as shown in Figure \ref{adjacency} (c) and (d). Finally, in case of \textit{unclumping}, a large number of road users are bunched together at the traffic signal that leads to a lot of interactions between the road users and is reflected as a dense adjacency matrix. As the traffic disperses, the road users move away from each other and the interactions become less pronounced that leads to sparser adjacency matrix as shown in Figure \ref{adjacency} (e) and (f). We observed that these three states broadly cover the behavior of all road users in mixed lane-less traffic at intersections. In case of highway traffic, neutral state is normal traffic behavior without congestion. The clumping state is the beginning of a congestion and the unclumping state reveals the end of congestion. Hence, all major traffic flow based predictions can be made by observing for these traffic states. 

\subsection{Spatio-temporal learning of traffic states}
As the adjacency matrices vary in density as well as over time, it is important to capture the spatial and temporal characteristics. To extract the spatial features, the adjacency matrices are passed through a CNN architecture and an output feature vector is obtained that represents the adjacency matrix at each time-step. 

The temporal relationship between the adjacency matrices is learnt using a network of Gated Recurrent Units (GRU) \cite{gru}. The concept of GRUs was introduced in \cite{gru} to solve the issue of vanishing gradient in Recurrent Neural Networks (RNN) by adding two gates - an update gate and a reset gate. The update gate $z_t$ for time step $t$ is computed as
\begin{equation}
    z_t = \sigma(\mathbf{W}^z\mathbf{x}_t + \mathbf{U}^z h_{t-1}),
\end{equation}
where $x_t$ is the input feature vector obtained from the adjacency matrix and is multiplied by the weight $W^z$. Similarly, $h_{t-1}$ holds the information for the previous $t-1$ time-steps and is multiplied by the weight $U^z$. The results of both these multiplications is added and passed through a sigmoid activation function to obtain an output between 0 and 1. The update gate helps the GRU in determining the amount of information from previous time-steps that needs to be passed to the future. This mechanism allows the GRU to retain gradients from the past and combats the vanishing gradient problem. For our use case, the change in adjacency matrices happen over a long time of around 50 time-steps. The update gate allows the GRU to retain the information about the changes over time.

The reset gate is used to determine the amount of past information that the GRU unit should forget and is given as
\begin{equation}
    r_t = \sigma(\mathbf{W}^r\mathbf{x}_t + \mathbf{U}^r h_{t-1}),
\end{equation}
where $\mathbf{W}^r$ and $\mathbf{U}^r$ are the weights associated with the reset gate. The way the reset gate is calculated is similar to the update gate. However, the usage of the reset gate is on the current memory content $h'_t$. The current memory content uses the reset gate to store relevant information from the previous time-steps and is calculated as 
\begin{equation}
    h'_t = \tanh(\mathbf{W}\mathbf{x}_t + r_t \odot \mathbf{U} h_{t-1}),
\end{equation}
where the input $\mathbf{x}_t$ is multiplied by a weight $\mathbf{W}$ and the previous state information $h_{t-1}$ with another weight $\mathbf{U}$. The reset gate $r_t$ is then multiplied element-wise (denoted by $\odot$) to the $\mathbf{U} h_{t-1}$ to remove appropriate information from the previous time steps. In case of the neutral state, no significant change occurs in the density of adjacency matrices then the reset gate can help the GRU retain only relevant information. 

Finally, the GRU unit needs to $h_t$ that holds the information for the current unit for passing down to the other units. The update gate is needed to determine the information to collect from the current memory content — $h’_t$ and the previous time-steps — $h_{t-1}$. The final memory content is computed as 
\begin{equation}
    h_t = z_t \odot h'_t + (1-z_t) \odot h_{t-1}.
\end{equation}
In the proposed approach, we employ 2 layers of GRU units before classifying the adjacency matrix sequence into the aforementioned 3 traffic states using a $softmax$ layer. 

\section{Experiments}\label{experiments}
The approach discussed in the previous section needs to be validated on some real-life mixed lane-less traffic data. The EyeonTraffic (EoT) dataset is curated for the same purpose.

\subsection{EyeonTraffic Dataset}
A total of 3 intersections were chosen for the EoT dataset with around 1 hour of aerial video recorded for each of the intersections, namely, \textit{Paldi (P), Nehru bridge - Ashram road (N)}, and \textit{APMC market (A)} in the city of Ahmedabad, India. These intersections were considered because of the diverse traffic conditions they present. While  \textit{Paldi} and \textit{Nehru bridge} are four-way signalized intersections, \textit{APMC market} is a three-way non-signalized intersection. Hence, this dataset comprehensively covers a wide variety of traffic conditions for both signalized and non-signalized intersections. The videos were captured using the included camera in the DJI Phantom 4 Pro drone at 50 frames per second in 4K resolution (4096$\times$2160). Using the RetinaNet architecture \cite{retinanet} for road user detection and the DeepSort tracker \cite{deepsort} for multi-object tracking, the road users of the entire intersection were detected and tracked. A total of 299,452, 294,769, and 202,433‬‬ trajectories were extracted for intersections P, N, and A, respectively.

\begin{figure}
    \centering
    \includegraphics[width=0.9\linewidth]{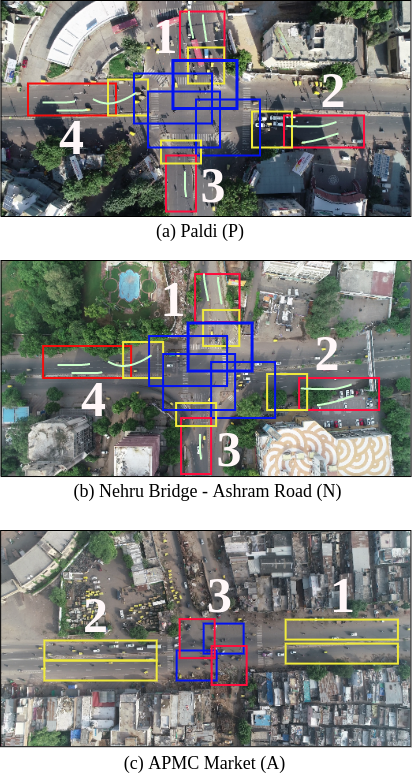}
    \caption{Spatial regions considered for annotation at each of the three intersections in the EoT dataset. Red denotes clumping, yellow denotes neutral, and blue denotes unclumping. Each way is denoted by a number.}
    \label{annotations}
\end{figure}

The 3 intersections were annotated by marking the spatial coordinates on one of the frames using LabelImg \cite{labelimg}, a graphical image annotation tool. As intersections P and N are 4 way intersections, each direction was individually labeled as shown in Figure \ref{annotations} (clump - red, neutral - yellow, unclump - blue). In total, a total of 12 spatial regions were found that represent the 3 traffic states in each of the 4 directions. Regarding intersection A that is a 3-way intersection, the left and right lanes exhibit neutral behavior and round-about exhibits both clumping and unclumping behavior resulting in a total of 8 spatial regions. 

In order to temporally annotate the videos where a particular traffic state was observed, the 3 hours of videos were divided into corresponding time intervals. Each time interval is denoted by start time, end time, and the traffic state. As shown in Figure \ref{annotations}, 1, 2, 3, and 4 are the direction codes and u, c, and n are the traffic state codes to make the processing easier. For example, 1u correspond to unclumping traffic state in direction 1. Next, for each annotated spatio-temporal region, the road users tracked within that region during the particular time interval are extracted. The extracted tracks are sampled at 5 frames per second and any spatio-temporal region with less than 20 unique road users is removed. Finally, each spatio-temporal region produces a sequence of road users that occupy that region. The total number of sequence for each traffic state obtained from the 3 intersections are shown in Table \ref{eotstats}.

\begin{table*}
\centering
\caption{Details of the EoT dataset.}
\label{eotstats}
\begin{tabular}{|c|c|c|c|c|c|c|c|}
\hline
\multicolumn{1}{|l|}{\multirow{2}{*}{\textbf{Intersection}}} &
  \multicolumn{3}{c|}{\textbf{Number of sequences}} &
  \multirow{2}{*}{\textbf{\begin{tabular}[c]{@{}c@{}}Min. no. of unique \\ road users/sequence\end{tabular}}} &
  \multicolumn{1}{l|}{\multirow{2}{*}{\textbf{\begin{tabular}[c]{@{}l@{}}Max. no. of unique \\ road users/sequence\end{tabular}}}} &
  \multicolumn{1}{l|}{\multirow{2}{*}{\textbf{\begin{tabular}[c]{@{}l@{}}Avg. duration of \\ sequence (in s)\end{tabular}}}} &
  \multirow{2}{*}{\textbf{\begin{tabular}[c]{@{}c@{}}Total\\ sequences\end{tabular}}} \\ \cline{2-4}
\multicolumn{1}{|l|}{} & \textbf{Clumping} & \textbf{Neutral} & \textbf{Unclumping} &    & \multicolumn{1}{l|}{} & \multicolumn{1}{l|}{} &     \\ \hline
P                      & 152               & 220              & 189                 & 41 & 97                    & 10                    & 561 \\ \hline
N                      & 79                & 100              & 115                 & 31 & 106                   & 8                     & 294 \\ \hline
A                      & 169                  & 141                 & 138                    &20    &90                       & 10                      & 448    \\ \hline
\end{tabular}
\end{table*}

\subsection{Experiment Settings}
The tracks obtained for each of the spatio-temporal region are used to create a corresponding adjacency matrix based on the road user ids. The distance between two road users is converted into metres from pixel values. If the distance is less than $\mu$= 10m (Equation \ref{equ}), the corresponding entry is added to the adjacency matrix based on road width. The image representation of the adjacency matrices is sent as input to VGG16 \cite{vgg} CNN architecture pretrained on ImageNet dataset. The input image is resized to 224$\times$224 and a 147 dimension feature vector is extracted from the average pool layer. VGG16 is a lightweight architecture that is capable of extracting the density changes in the adjacency matrices. In order to normalize the length of adjacency matrix sequences, we fixed the number of frames per sequence to 50 based on the minimum median value across the 3 interactions in the EoT dataset. Hence, the sequences for each intersection can be represented as 3D tensors of size $n_s \times 50 \times 147 $ where $n_s$ is the number of sequences for that intersection. Finally, the sequences are divided randomly into 70\% training (910 sequences), 10\% validation (132 sequences), and 20\% testing (261 sequences). 

\subsection{Comparison of Temporal Networks}
To learn the temporal structure in the sequences, a variety of temporal networks were tested as listed below:
\begin{itemize}
\item GRU(100,50): 2 GRU layers with 100 and 50 units followed with a dense layer with 30 units with Rectified Linear Unit (ReLU) activation
\item GRU(50,25): 2 GRU layers with 50 and 25 units
\item GRU-A(100,50): 2 GRU layers with 100 and 50 units followed by an attention layer and dense layer with 30 units with ReLU activation
\item LSTM(100,50):  2 LSTM layers with 100 and 50 units 
\item LSTM-A(100,50):  2 LSTM layers with 100 and 50 units followed by an attention layer and dense layer of 30 units with ReLU activation 
\item RNN(100,50): 2 vanilla RNN layers with 100 and 50 units
\item RNN-A(100,50):  2 vanilla RNN layers with 100 and 50 units  followed by an attention layer and dense layer of 30 units with ReLU activation
\end{itemize}
For all the temporal networks described above, the categorical crossentropy loss function is used for classification and the networks are trained with Adam optimizer. After hyper-parameter optimization, the most suitable values for learning rate, number of epochs, batch size, and recurrent dropout were found to be 0.001, 300, 32, and 0.6. 

The results of the aforementioned networks are presented in Table \ref{comparison}. It can be observed that the GRU(100,50) obtains the best classification accuracy across all the temporal networks. This shows that GRU can adequately capture the temporal density changes in the adjacency matrices. Further, the sequences do not have enough complex temporal dependencies that an LSTM network is required. Hence, the large number of parameters in the LSTM network cannot be trained properly given the sequence data. This is also the reason that vanilla RNN has comparable performance to LSTM. Adding attention to the networks causes an improvement in the classification performance. Hence, some parts of the sequence is crucial to understanding the traffic state rather than the entire sequence. Especially, the rapid movement of vehicles towards the end of clumping or beginning of unclumping are essential in identify the corresponding traffic states. 

\begin{table}[]
\caption{Comparison of various temporal networks}
\label{comparison}
\begin{tabular}{|c|c|c|c|c|}
\hline
\multirow{2}{*}{\textbf{Temporal network}} & \multicolumn{4}{c|}{\textbf{Accuracy (\%)}}      \\ \cline{2-5} & \textbf{neutral} & \textbf{clumping} & \textbf{unclumping} & \textbf{Total} \\ \hline
GRU(100,50) & \textbf{69.89}                     & 51.31                   & 68.47                     & 64                           \\ \hline
GRU(50,25) & 62.36                     & 50.0                    & 69.56                     & 61.3                         \\ \hline
GRU-A(100,50)           & 65.59                     & 40.78                   & \textbf{82.60}                     & \textbf{64.4}                         \\ \hline
LSTM (100,50)       & 62.36                     & 47.36                   & 71.73                     & 61.3                         \\ \hline
LSTM-A(100,50) & 64.51                     & \textbf{60.52}                   & 66.30                     & 64                           \\ \hline
RNN(100,50) & 67.74                     & 28.94                   & 70.65                     & 57.5                         \\ \hline
RNN-A(100,50)  & 63.44                     & 43.42                   & 73.91                     & 61.3                         \\ \hline
\end{tabular}
\end{table}

Table \ref{comparison} also presents the classification accuracy for each of the traffic states separately. It can be seen that classification accuracy is high for unclumping and neutral states compared to clumping.  It shows that both unclumping and neutral have more distinct patterns with low intra-class variability. In case of clumping, the presence of many stationary vehicles near the traffic signal causes misclassification with the neutral state. The evidence for this can be found when the analysis is done for every intersection.

\begin{figure}
    \begin{tabular}{ccc}
    \includegraphics[width=0.3\linewidth]{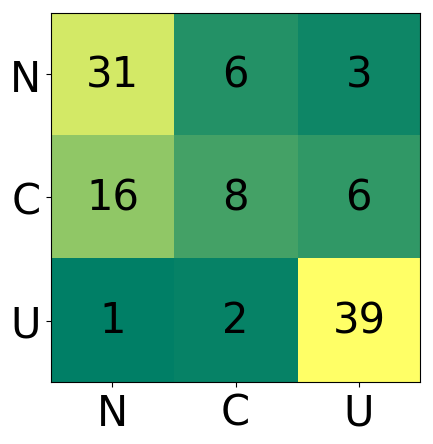}&
    \includegraphics[width=0.3\linewidth]{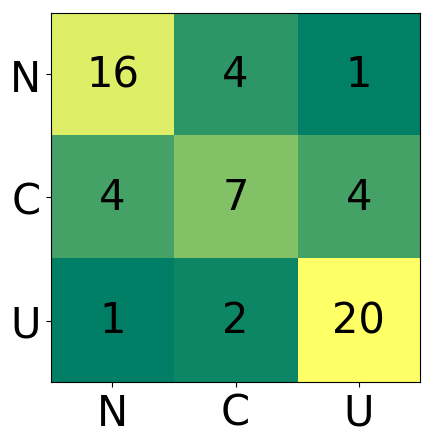}&
    \includegraphics[width=0.3\linewidth]{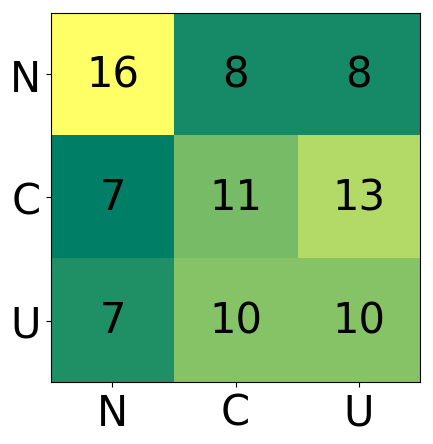} \\
    (a) Intersection P & (b) Intersection N & (c) Intersection A
    \end{tabular}
    \caption{Intersection-wise comparison of traffic state prediction using the best performing GRU-A(100,50) network. N - neutral, C - clumping, U - unclumping.}
    \label{confusion}
    \vspace{-5mm}
\end{figure}

It is also important to analyze the results for every intersection separately as the traffic flow behaviors are different. In Figure \ref{confusion}, we present the confusion matrices for each intersection separately using the best performing GRU-A(100,50) network. As stated earlier, for all the intersections, unclumping is mostly misclassified as neutral. Furthermore, the traffic states are better predicted in case of intersection P and N (69.6\% and 86.0\%, respectively) compared to A (41.1\%). As intersection A is an unsignalized intersection, the traffic states like unclump and clump are not as pronounced.  This leads to higher misclassification as compared to other intersections where the traffic signal brings out a more pronounced behavior.

\setlength{\tabcolsep}{2pt}

\subsection{Leave-one-out testing}
\begin{figure}
    \begin{tabular}{|c|c|c|}
    \hline
    \multicolumn{3}{|c|}{Tested intersection} \\
    \hline
    \textbf{Intersection P }& \textbf{Intersection N} &  \textbf{Intersection A} \\
    \hline
    \hline
    \includegraphics[width=0.3\linewidth]{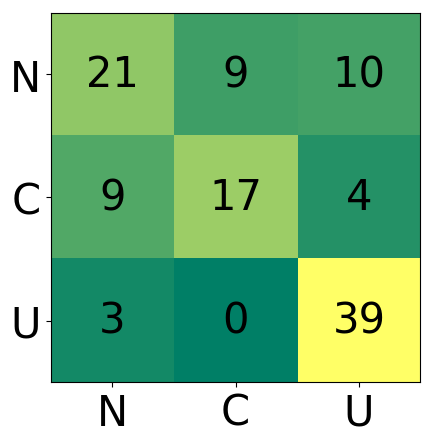}&
    \includegraphics[width=0.3\linewidth]{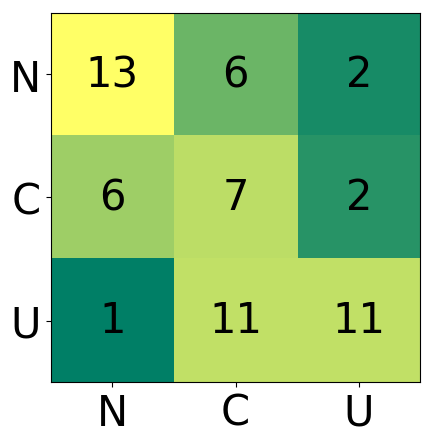}&
    \includegraphics[width=0.3\linewidth]{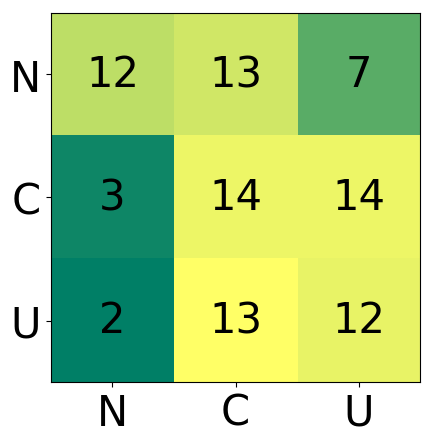} \\
    \multicolumn{3}{|c|}{ \textbf{(a) Trained on Intersection P}} \\
    \hline
    \hline
    \includegraphics[width=0.3\linewidth]{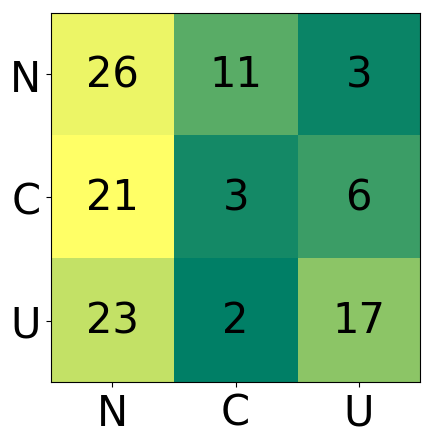}&
    \includegraphics[width=0.3\linewidth]{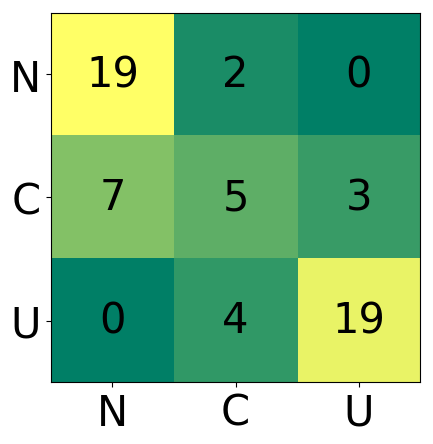}&
    \includegraphics[width=0.3\linewidth]{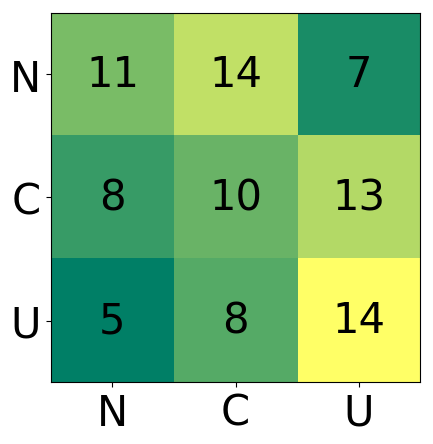} \\
    \multicolumn{3}{|c|}{\textbf{(b) Trained on Intersection N}}\\
    \hline
    \hline
    \includegraphics[width=0.3\linewidth]{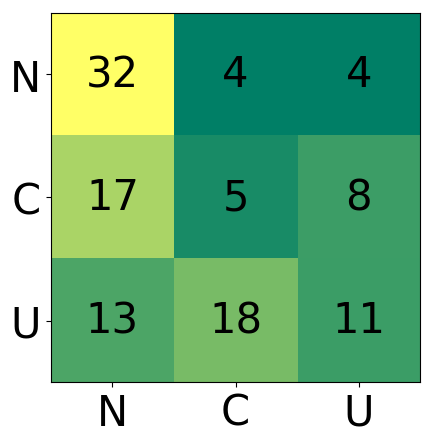}&
    \includegraphics[width=0.3\linewidth]{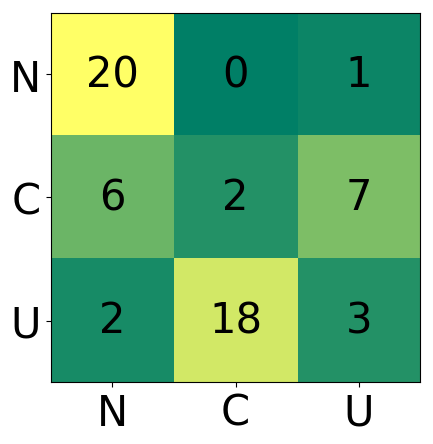}&
    \includegraphics[width=0.3\linewidth]{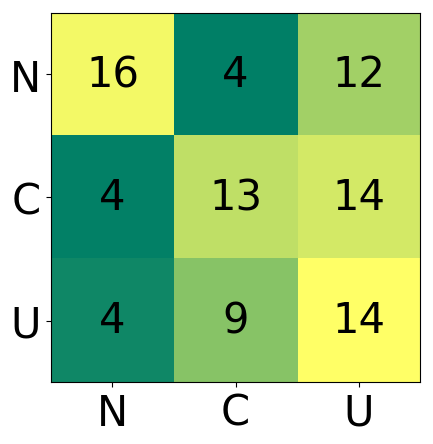} \\
    \multicolumn{3}{|c|}{\textbf{(c) Trained on Intersection A}} \\ \hline
    \end{tabular}
    \caption{Prediction performance of various intersections when trained in leave-one-intersection-out fashion. Network used is GRU-A(100,50).}
    \label{leaveoneout}
\end{figure}

In order to understand the generalization ability of the proposed approach, we devise an experiment where the temporal network is trained with sequences from one intersection and tested on all three intersections. This mimics actual scenarios where labelled sequences are not available for a new intersection for which we have to predict the traffic states. In Figure \ref{leaveoneout}, we present the results when the training data of different intersections is used to train the best performing temporal network GRU-A(100,50). It can be seen that the temporal network generalizes well when trained on intersection A yielding 40.17\%, 42.37\%, and 47.77\% for intersections P, N, and A, respectively. However, when trained on either P or N, the prediction performance does not translate well to the other intersections. Hence, the generalization performance is not as robust across intersections with varying lane configurations.

\section{Conclusion}
In this work, an approach to identify traffic states in lane-less traffic using temporal changes in the traffic graph was proposed. We showed that a spatio-temporal CNN-GRU network applied on the adjacency matrix of a traffic graph can identify clumping, unclumping, and neutral traffic states in various spatial regions of an intersection. Further, we showed the effectiveness of the proposed approach on a large annotated aerial dataset called EyeonTraffic that covered 3 intersections in Ahmedabad, India. In future, we would like to apply this approach to predict the onset of a congestion by observing the changes in traffic behavior.

\section*{Acknowledgement}
This work has been conducted as the part of SATREPS project entitled on “Smart Cities for Emerging Countries by Multimodal Transport System based on Sensing, Network and Big Data Analysis of Regional Transportation” (JPMJSA1606) funded by JST and JICA. 

\bibliographystyle{IEEEtran}
\bibliography{main.bib}

\end{document}